\documentclass[letterpaper, 10 pt, conference]{ieeeconf}  

\IEEEoverridecommandlockouts                              

\usepackage[utf8]{inputenc}
\usepackage{pgfplots}
\DeclareUnicodeCharacter{2212}{−}
\usepgfplotslibrary{groupplots,dateplot}
\usetikzlibrary{patterns,shapes.arrows}
\usetikzlibrary{fadings}
\pgfplotsset{compat=newest}

\usepackage{times}

\usepackage{multicol}
\usepackage[bookmarks=true]{hyperref}
\usepackage{amsmath,amssymb,amsfonts}
\usepackage{booktabs}
\usepackage[font=footnotesize]{caption}
\usepackage{graphicx}
\usepackage{xcolor}
\usepackage{bbm}
\usepackage{nicefrac}
\usepackage{comment}
\usepackage{algorithm}
\usepackage{algpseudocode}
\usepackage{hyperref}

\newcommand\argmin[1]{\underset{#1}{\text{argmin }}}
\newcommand\argmax[1]{\underset{#1}{\text{argmax }}}

\newcommand{\sref}[1]{Sec. \ref{#1}}
\newcommand{\figref}[1]{Fig. \ref{#1}}
\newcommand{\tabref}[1]{Table \ref{#1}}

\newcommand{\NN}{\mathcal{N}}

\newtheorem{theorem}{Theorem}[section]

\newtheorem{remark}[theorem]{Remark}

\title{\LARGE \bf
Towards Proactive Safe Human-Robot Collaborations via \\Data-Efficient Conditional Behavior Prediction
}

\author{Ravi Pandya$^{*,1}$, Zhuoyuan Wang$^{*,2}$, Yorie Nakahira$^2$, Changliu Liu$^1$
\thanks{*contributed equally to this work.}
\thanks{$^1$Authors are with the Robotics Institute at Carnegie Mellon University, Pittsburgh, Pennsylvania, \tt\small rapandya, cliu6@andrew.cmu.edu}
\thanks{$^2$Authors are with the Department of Electrical and Computer Engineering at Carnegie Mellon University, Pittsburgh, Pennsylvania, \tt\small zhuoyuaw, ynakahir@andrew.cmu.edu}
}

\begin{document}

\maketitle
\thispagestyle{plain}
\pagestyle{plain}

\begin{abstract}
We focus on the problem of how we can enable a robot to collaborate seamlessly with a human partner, specifically in scenarios 
where preexisting data is sparse. Much prior work in human-robot collaboration uses \textit{observational} models of humans (i.e. models that treat the robot purely as an \textit{observer}) 
to choose the robot's behavior, but such models do not account for the influence the robot has on the human's actions, which may lead to inefficient interactions. We instead formulate the problem of optimally choosing a collaborative robot's behavior based on a \textit{conditional} model of the human that depends on the robot's future behavior. First, we propose a novel model-based formulation of conditional behavior prediction that allows the robot to infer the human's intentions based on its future plan in data-sparse environments. We then show how to utilize a conditional model for proactive goal selection and safe trajectory generation around human collaborators. Finally, we use our proposed proactive controller in a collaborative task with real users to show that it can improve users' interactions with a robot collaborator quantitatively and qualitatively.

\end{abstract}

\IEEEpeerreviewmaketitle

\section{Introduction}
As robots are becoming more common in industrial manufacturing, social, and home environments, it is imperative that they seamlessly collaborate with humans. 
This would allow us to take advantage of the speed and precision of robots as well as the flexibility of humans for completing tasks (such as in flexible industrial production lines~\cite{kruger2009cooperation, villani2018survey}). Instead of passively reacting to the human's behavior, we also want robots to \textit{proactively} collaborate with humans by estimating and influencing their future behavior. One major challenge  is the closed-loop nature of the interaction---the robot needs to understand what the human wants to do so it can assist them, but the robot's own actions will also influence the human's actions. 
Reasoning about this influence is an important factor in both efficient and safe human-robot interaction (HRI). 
Much existing work in the field of HRI has made use of observational models of humans~\cite{baker2007goal, baker2009action}, but generally lacks the ability to predict how the human will respond to the robot's ultimate decision. Game-theoretic approaches do this reasoning, but often assume some hierarchical information structure~\cite{fisac2019hierarchical}, can be too restrictive to achieve optimal safety-performance tradeoffs in human-machine systems~\cite{zhang2023rethinking}, and may be poor predictors of humans~\cite{bornhorst2004people, bruttel2012infinity, gachter2004behavioral}. 

Autonomous driving researchers have taken note of the importance of influence and recently developed predictive models of agents that are conditioned on the ego agent's future plan~\cite{tolstaya2021identifying}, sometimes called \textit{conditional behavior prediction} (CBP) models. CBP models rely on large datasets of human-human or human-robot interactions to reason about how a robot's actions would change the intentions of humans. 
Such datasets, however, are not available in many human-robot collaboration domains such as industrial manufacturing and home robotics. In particular, these environments have a variety different task specifications, so it may be infeasible to collect enough data for all tasks of interest. As a result, existing CBP techniques cannot be directly applied.


\begin{figure}
    \centering
    \includegraphics[width=\columnwidth]{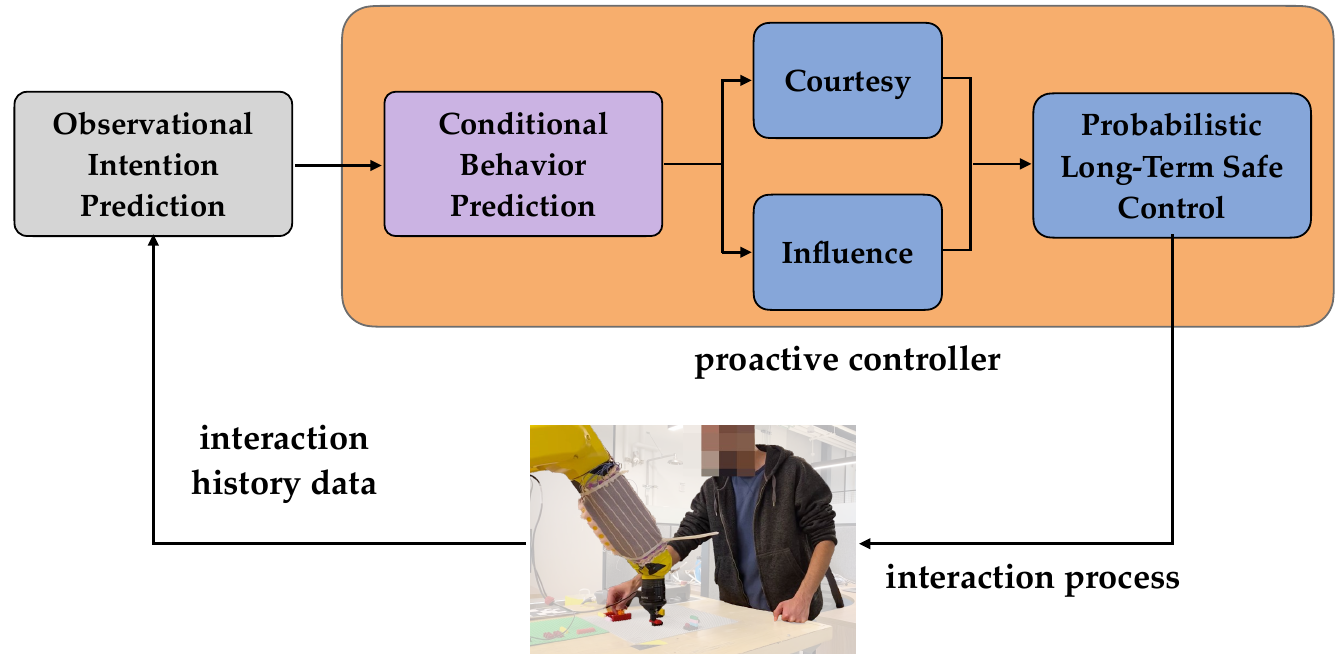}
    \caption{Overall diagram of the proposed framework, including the CBP model (\sref{sec:cbp_model}), courtesy and influence behavior (\sref{sec:proposed_controller}) and long-term safe control (\sref{sec:safety}).}
    \label{fig:overall_diagram}
    \vspace{-0.25in}
\end{figure}

In this paper, we apply the CBP idea more broadly to HRI across data-sparse domains. We take a step towards enabling robots to similarly reason about the closed-loop collaboration with human partners in a data-efficient manner using a model-based CBP formulation. We then utilize this CBP model for downstream collaborative tasks by introducing an objective that proactively 
switches between courteous and influential behavior. We additionally show how to utilize a CBP model for long-term safe control in multi-goal human-robot collaborations. The system diagram of the proposed framework is shown in \figref{fig:overall_diagram}.
Our contributions in this work are the following:
\begin{enumerate}
    \item Model-based conditional behavior prediction (CBP) in multi-goal human-robot collaborations.
    \item An integrated framework for courtesy and influence behavior in robot goal selection utilizing a CBP model.
\end{enumerate}

The merits of the proposed framework are validated using human experiments on multi-goal collaboration tasks.



\section{Related Work}

\noindent\textbf{Human Goal Prediction:} Much prior work has been done that treats humans as rational agents and tries to predict their intentions~\cite{baker2007goal, baker2009action}, often using the framework of inverse reinforcement learning~\cite{ng2000algorithms} and Bayesian inference~\cite{bestick2018learning, losey2019enabling}. Generally this line of work treats the robot as a \textit{passive observer} of the human's behavior~\cite{vasquez2008intentional} and has the robot pick a best response to the prediction~\cite{fisac2018probabilistically, mainprice2013human}. Some work has considered the case where both agents have goals in the same space to complete a collaborative task~\cite{pellegrinelli2016human, levine2018watching, gombolay2015decision}, but still do not explicitly condition on the robot's future actions.

\noindent\textbf{Conditional Behavior Prediction:} Autonomous agents influencing humans has been considered recently in autonomous driving and pedestrian prediction settings~\cite{tolstaya2021identifying, salzmann2020trajectron++}. These CBP approaches have been applied in scenarios where there are existing large datasets that allow large models to be trained~\cite{rhinehart2019precog, schmerling2018multimodal, tang2022interventional} (researchers in~\cite{tolstaya2021identifying} note that their dataset consists of 18 years of continuous driving data). We are instead interested in general HRI where data may be limited, so we formulate a CBP approach that allows the robot to reason in other data-sparse environments.

\noindent\textbf{Courtesy and Influence:} Prior work on courtesy and influence behavior for robots often relies on human and robot collision avoidance~\cite{yu2021path, sadigh2016planning} or minimizes the expected cost incurred by a human~\cite{sun2018courteous, bestick2017implicitly}. Other work on influencing humans usually rely on learning-based models, which are shown to be brittle with real humans~\cite{xie2021learning, wang2022influencing}.











\noindent\textbf{Safe Human-Robot Collaboration (HRC):}
Safety has long been studied for HRC with techniques like workplace separation~\cite{robla2017working}, rule-based methods~\cite{zanchettin2015safety} and control barrier functions~\cite{ferraguti2020safety, liu2015safe}. These techniques ensure safety reactively, so are only safe in the short-term. We are interested in ensuring long-term safety for HRC, so we leverage a long-term safe control strategy for stochastic systems~\cite{wang2022myopically} and discuss how to incorporate our CBP model into the safe controller.

\section{Problem Formulation}
\noindent\textbf{The HRI System.} We consider the case of one robot collaborating with one human. The two agents have a shared set of goals $\Theta$, known to both agents a priori. The human has noisy dynamics that depend on both agents' previous states and the human's true goal, denoted by $\theta_H^{*}$:
\begin{equation}
    x_H^{t+1} = m(x_H^t,x_R^t,\theta_H^{*}) + w^t,
\end{equation}
where $w^t$ is zero-mean Gaussian noise $w^t\sim\NN(0,\Sigma_H)$ with covariance $\Sigma_H$. 
The robot has deterministic dynamics: 
\begin{equation}
    x_R^{t+1} = f_R(x_R^t, u_R^t),
\end{equation}
where the robot's control policy $u_R^t$ may arbitrarily depend on the two agents states $x_H^t, x_R^t$ and the set of possible goals $\Theta$. We assume that the robot has access to the form of the human's dynamics function $m(x_H^t,x_R^t,\theta)$, but does not know the actual goal of the human $\theta_H^{*}$ or how it might change over time. The human's dynamics may be estimated offline from data~\cite{liu2020human} or from online adaptation~\cite{pandya2022safe}, but our focus is on dealing with uncertainty in the human's goal selection (particularly how it is affected by the robot).

\noindent\textbf{The Robot's Objective.} The robot's objective is to allow the team to work together fluently while keeping the human safe. The robot keeps a probabilisitic predictive model of the human $p(X_H\mid X_R=\mathbf{x_R}, \mathbf{o})$, where the future trajectories of the human and robot agents are random variables denoted by $X_H$ and $X_R$ respectively, and $\mathbf{o}$ contains observations of the environment and $\mathbf{x_R}=[x_R^t,...,x_R^{t+T}]$ is a particular robot future trajectory (similarly $\mathbf{x_H}$ for human). Since the robot's prediction is probabilistic, it needs to stay safe probabilistically. We thus want the probability of collision with the human to be less than a small value $\epsilon$, meaning the robot needs to stay safe with probability $1-\epsilon$. Based on a predefined safety specification, we define the set of safe states $\mathcal{Z}$. Let the joint state $x^t=[x_H^{t,\top}, x_R^{t,\top}]^\top$, so our probabilistic safety constraint is: 
\begin{equation}
\label{eq:safety_objective}
    p(x^{t}\in\mathcal{Z}, \; \forall t={1,..., H})\geq 1-\epsilon,
\end{equation}
where $t$ is the timestep and $H$ is the trajectory horizon. 

Finally, we are given some objective $J$ for the robot to minimize. This objective function could encompass both efficiency of completing the robot's own task and a penalty for affecting the human's intention. We investigate the design of $J$ in \sref{sec:proposed_controller}. Our robot's constrained objective is:
\begin{subequations}
\begin{align}
    \min_{\mathbf{x_R}} &\quad\mathbb{E}_{\mathbf{x_H}\sim p(X_H\mid X_R=\mathbf{x_R}, \mathbf{o})} \sum_{t=1}^H J(x_R^t, x_H^t) \\
    &\text{s.t.}\quad  
    p(x^{t}\in\mathcal{Z}, \; \forall t={1,..., H})\geq 1-\epsilon.
\end{align}
\end{subequations}

\begin{figure}
    \centering
    \includegraphics[width=0.9\columnwidth]{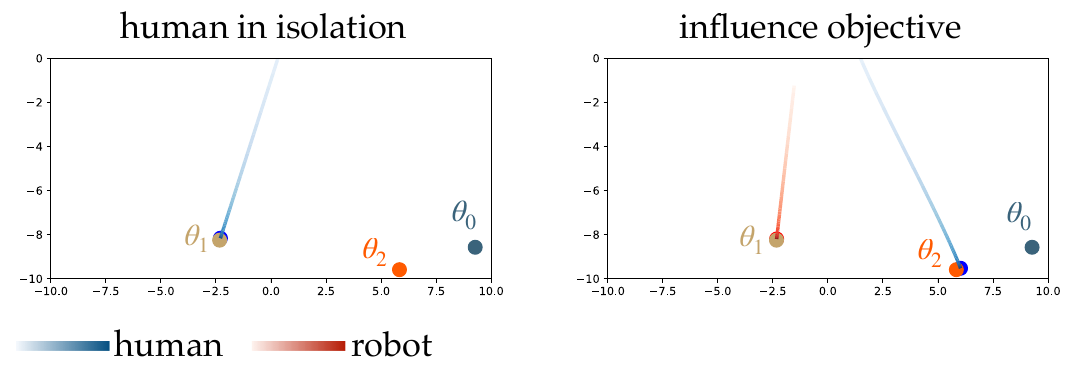}
    \caption{\textbf{Left:} human chooses left goal in isolation. \textbf{Right:} robot chooses a goal to successfully influence the human's goal selection by modeling their goal change using model-based CBP.}
    \label{fig:influence_obj}
    \vspace{-0.2in}
\end{figure}

\label{sec:sim_env}
\noindent\textbf{Simulation Environment.} In all simulations, the human and the robot are double-integrator LTI systems:
\begin{align}
    x_H^{t+1} &= m(x_H^t, x_R^t, \theta_H^*) + w^t= Ax_H^t + Bu_H^t + w^t\\
    x_R^{t+1} &= f_R(x_R^t, u_R^t) = Ax_R^t + Bu_R^t
\end{align}
where 
$x_{\{H,R\}}=[p_x, v_x, p_y, v_y]^\top$ and 
$w^t\sim \mathcal{N}(0,\Sigma_H)$. The  human's control is inspired by social force models of humans~\cite{helbing1995social}, where they are attracted to their goal $\theta_H^*$ and repelled from the robot:
\begin{equation}
    u_H^t = K(\theta_H^* - x_H^t) + \frac{\gamma}{d^2}(C_H x_H^t - C_R x_R^t).
\end{equation}
Here, $K$ is the control gain computed from LQR, $C_H$ and $C_R$ are  matrices that select the $(x,y)$ positions of agents' states, $d = || C_R x_R^t - C_H x_H^t ||_2$ is the distance between the two agents,
and $\gamma$ controls how strongly the human is repelled from the robot. The two agents are trying to reach goal states from a set of three fixed goals without colliding with each other. The simulation environment can be seen in \figref{fig:influence_obj}.


\section{Model-Based CBP}
\label{sec:cbp_model}

\noindent\textbf{Key Insight.} As previously noted, prior work using observational inference over a human's goal does not account for potential effects of a robot collaborator, since the robot is treated as an external observer. Our key insight is that we can \textit{decouple} the human's \textit{prior} goal $\theta_H^{prior}$ (from observational inference) from their \textit{posterior} goal they will ultimately choose around the robot, $\theta_H^{post}$. This allows us to formulate a Bayesian-inference-based CBP model that can be used for HRC without the need for large datasets.

\noindent\textbf{Derivation.}
Observational Bayesian inference gives the robot access to
\begin{equation}
    b^t_R(\theta_H^{prior}) := p(\theta_H^{prior} \mid x_H^{0:t}, x_R^{0:t}, u_H^{0:t}).
\end{equation}
Given the simulated human's LQR controller (\sref{sec:sim_env}), we follow~\cite{tian2023towards} to compute an exact form of this belief.

We want the robot to estimate the distribution of $\theta_H^{post}$ conditioned on the robot's future plan: $b^t_R(\theta_H^{post} \mid x_R^{t+1:T}) = p(\theta_H^{post} \mid x_H^{0:t}, x_R^{0:t}, u_H^{0:t}, x_R^{t+1:T})$. However, we know that it may be computationally intractable to integrate over the space of future trajectories. We instead consider the robot's goal $\theta_R$ as a proxy for the robot's full plan:
\begin{equation}
    b^t_R(\theta_H^{post} \mid \theta_R) := p(\theta_H^{post} \mid x_H^{0:t}, x_R^{0:t}, u_H^{0:t}, \theta_R).
\end{equation}
We note that one could include $\theta_R$ in the human's reward function in observational Bayesian inference directly. However, decomposing the inference as we propose frees the robot to convert any nominal intention predictor (even learning-based) to a conditional one. 

For ease of notation, we denote the robot's \textit{observation} $(x_H^{0:t}, x_R^{0:t}, u_H^{0:t})$ as $\mathbf{o^{0:t}}$. To compute the conditional distribution, we can integrate out the variable $\theta_H^{prior}$ from the joint distribution $p(\theta_H^{post}, \theta_H^{prior} \mid \mathbf{o}^{0:t}, \theta_R)$ to get
$p(\theta_H^{post} \mid \mathbf{o^{0:t}}, \theta_R) = \int_{\theta_H^{prior}} p(\theta_H^{post} \mid \theta_H^{prior}, \mathbf{o^{0:t}}, \theta_R) p(\theta_H^{prior} \mid \mathbf{o^{0:t}}, \theta_R)$.
We know that the human's \textit{prior} goal selection $\theta_H^{prior}$ will be independent of the robot's \textit{future} goal $\theta_R$, so we can simplify this equation as
$ \int_{\theta_H^{prior}} p(\theta_H^{post} \mid \theta_H^{prior}, \mathbf{o^{0:t}}, \theta_R) p(\theta_H^{prior} \mid \mathbf{o^{0:t}})
$. 
Since we have a discrete set of goal locations, the final formula is
\begin{equation}
    \begin{split}
    &p(\theta_H^{post} \mid \mathbf{o^{0:t}}, \theta_R)\\
    &= \sum_{\theta_H^{prior}} p(\theta_H^{post} \mid \theta_H^{prior}, \mathbf{o^{0:t}}, \theta_R) p(\theta_H^{prior} \mid \mathbf{o^{0:t}}).
\end{split}
\end{equation}
Note that the second term is exactly what comes out of observational Bayesian goal inference. Now we need a way to compute $p(\theta_H^{post} \mid \theta_H^{prior}, \mathbf{o^{0:t}}, \theta_R)$.
Inspired by prior work~\cite{dragan2013legibility, bestick2018learning}, we use a softmax distribution with well-designed features as a strong prior so we can utilize this formulation in data-sparse environments:
\begin{equation}
    p(\theta_H^{post} \mid \theta_H^{prior}, \mathbf{o^{0:t}}, \theta_R) = \frac{e^{-\beta_{cbp} s(\theta_H^{post}, \theta_R; \mathbf{o^{0:t}}, \theta_H^{prior})}}{\sum_{\theta_H} e^{-\beta_{cbp} s(\theta_H, \theta_R; \mathbf{o^{0:t}}, \theta_H^{prior})}},
\end{equation}
where $\beta_{cbp}$ is an additional inverse temperature parameter for this distribution and $s(\cdot)$ is a score function that encodes how much robot's goal selection influences the human's goal. 


\section{The Robot's Multi-Stage Objective}
\label{sec:proposed_controller}
\noindent\textbf{Design Goal.} Equipped with a CBP model, we wish to design an objective $ J(\cdot)$ that will enable efficient collaborations with a human partner. We first note that many collaborative tasks can be solved in a leader-follower manner~\cite{van2020adaptive, messeri2020effects} where one agent decides on a general strategy and the other follows along. When interacting with humans, we want robots to be \textit{courteous} so as to not interfere with the human's intended strategy. However, there may be times where the human is uncertain about what strategy to choose. At such times, we want the robot to \textit{proactively influence} the human to choose an efficient strategy. We introduce a switching controller that estimates whether the human is hesitating then utilizes its CBP model to be either proactive or courteous.


\begin{figure}
    \centering
    \includegraphics[width=0.9\columnwidth]{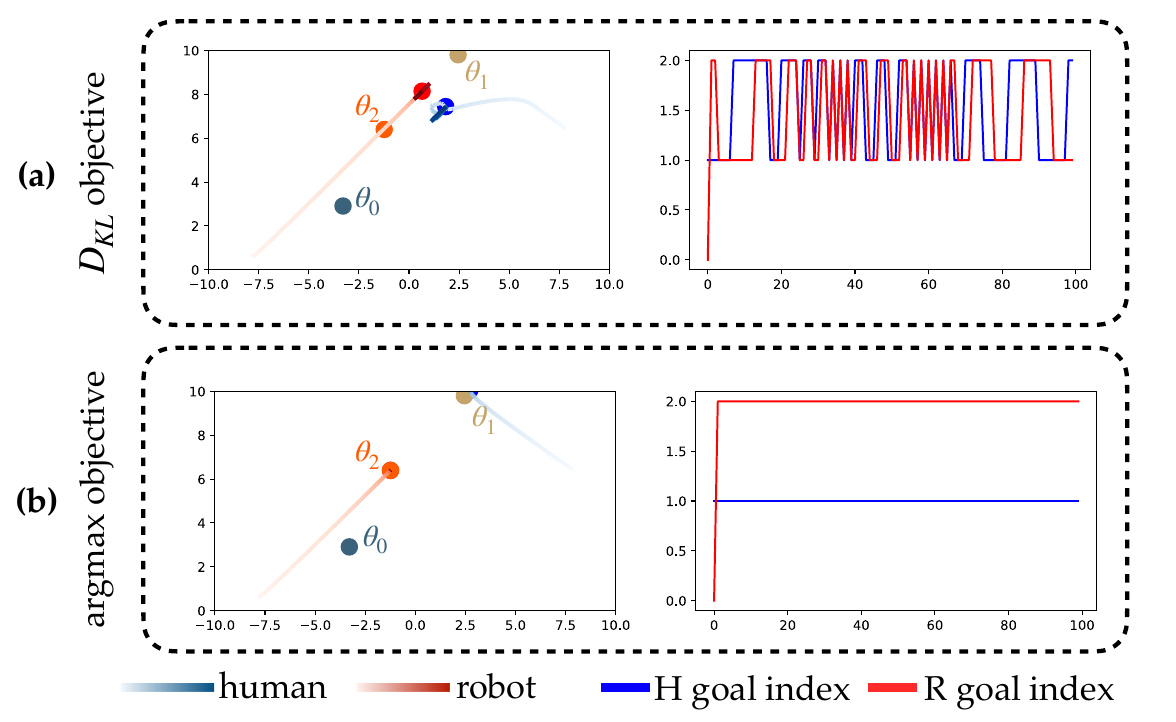}
    \caption{Visualization of interaction and goal selection with \textbf{(a)} KL-divergence cost function \textbf{(b)} argmax belief cost function. The KL-divergence cost results in chattering of both agents' goals while the argmax belief results in a stable interaction.}
    \label{fig:courtesy_obj}
    \vspace{-0.2in}
\end{figure}

\noindent\textbf{Detecting Human Uncertainty. }
We aim to equip the robot with the ability to detect when the human is uncertain of their goal and subsequently influence them towards an efficient strategy. 
To do this, we turn to prior work on detecting model misspecificsation in Bayesian models~\cite{fisac2018probabilistically, bobu2020quantifying}. We jointly keep a belief over the human's prior intention $\theta_H^{prior}$ and the robot's model-confidence parameter $\beta_R$:
\begin{equation}
    b_R^{t+1}(\theta_H^{prior},\beta_R) = \frac{p(u_H^t\mid \mathbf{o^{0:t}}; \theta_H^{prior}, \beta_R) b_R^t(\theta_H^{prior},\beta_R)}{\int_{\theta,\beta} p(u_H^t\mid \mathbf{o^{0:t}}; \theta, \beta) b_R^t(\theta,\beta) d\theta d\beta}.
\end{equation}
We follow prior work~\cite{bobu2020quantifying} to detect when the model confidence is low: $\forall\theta\in\Theta, \text{argmax}_{\beta} b_R(\beta \mid \theta) < \delta$.
If all hypotheses have the most probability mass on $\beta$s that are smaller than a threshhold $\delta$, the robot will raise a flag saying that the human is uncertain about their intention. Since we assume both agents know the set of goals, the most likely cause of low model confidence is that the human has not decided on their goal.



\noindent\textbf{Following Behavior: Courtesy Objective. }
Intuitively, avoiding changing the human's intention might be formulated as minimizing the KL-divergence between the robot's nominal belief of the human and the conditional belief of the human, which would mean that the robot's future actions have no effect on the human's intention:
\begin{equation}
    J(\theta_R) = D_{\text{KL}}\left(b^{t+1}_R(\theta_H^{post} \mid \theta_R) \parallel b_R^t(\theta_H^{prior})\right).
\end{equation}
We actually find that this approach results in chattering of the robot's goal (\figref{fig:courtesy_obj}(a)).
Since the robot is trying to keep the \textit{belief distribution} close to the prior belief, it changes its goal to keep its belief from changing over time. 
Instead, we consider maximizing the probability that the human keeps the same intention before and after considering the robot's action. The objective function to \textit{minimize} is:
\begin{equation}
\begin{split}
    J_c(\theta_R) &= -b_R^{t+1}(\theta_H^{post}=\hat\theta_H \mid \theta_R), \\ 
    \text{where } \hat\theta_H &= \argmax{\theta\in\Theta} b_R^t(\theta_H^{prior}=\theta).
\end{split}
\label{eqn:courtesy_obj}
\end{equation}
Using this courtesy objective results in stable goal selection for the robot, seen in \figref{fig:courtesy_obj}(b). 

\noindent\textbf{Leading Behavior: Influence Objective. }
When the human is uncertain of which goal to choose, the robot has a chance to influence them. We want the interaction to be efficient, so the robot can pick a goal for the human that it thinks is best for the team and try to influence the human to move towards it. The cost function is the total distance that the team would travel:
\begin{equation}
\begin{split}
    J_i(\theta_R) &= || x_R^t - \theta_R || + || x_H^t - \hat\theta_H ||, \\
    \text{where } \hat\theta_H &= \argmax{\theta\in\Theta} b_R^{t+1}(\theta_H^{post}=\theta \mid \theta_R).
\end{split}
\label{eqn:influence_obj}
\end{equation}
Using this objective function and the CBP model, the robot influences the simulated human to select a different goal than they would have in isolation (\figref{fig:influence_obj}).

\noindent\textbf{Overall Objective.} The robot's objective switches modes between \textit{influencing} the human and being \textit{courteous} to them based on whether the robot believes the human is unsure about what to do:
\begin{equation}
    J(\cdot) =
    \begin{cases}
      J_i(\cdot), & \text{if } \forall\theta\in\Theta, \text{argmax}_{\beta} b_R(\beta \mid \theta) < \delta, \\
      J_c(\cdot), & \text{otherwise}.
    \end{cases}
    \label{eqn:final_objective}
\end{equation}


\section{Safe Trajectory Generation}
\label{sec:safety}
We still need the robot to stay safe in the long term with respect to the actual \textit{conditional} model of the human, since this will be the distribution that the human's goal selection ultimately takes. One major challenge is that the uncertainty in the human's intention hinders the direct application of short-term-based safe control methods to ensure long-term safety~\cite{he2023hierarchical,wei2019safe}. We instead describe a method for allowing an existing \textit{long-term} probabilistic safe controller~\cite{wang2022myopically} to sample directly from the CBP model to compute safe trajectories.
We define safety as keeping a minimum distance $d_\text{min}$ between human and robot. Mathematically, we can define a barrier function 
\begin{equation}
    \phi_{d_\text{min}}(x) =|| C_R x_R - C_H x_H ||_2 - d_\text{min}
\end{equation}
The safe set of states for the system is then $\mathcal{Z} = \{x: \phi_{d_\text{min}}(x)\geq 0\}$. To use the CBP model (\sref{sec:cbp_model}) for safe control, the robot needs to know how its low-level action will affect the human's intention. 

This can be done by letting the robot keep a \textit{mental model} of the human's inference over the robot's goal $\hat b_H^t(\theta_R)$ (a common paradigm in the literature~\cite{dragan2013legibility, huang2019nonverbal}). The mental model is essentially an observational Bayesian inference model of the \textit{robot's} goal. The mental model can be simulated forward and used to keep an overall belief over the human's posterior goal $\theta_H^{post}$: $p(\theta_H^{post})=\hat b_H^t(\theta_R) b_R^t(\theta_H^{post}\mid \theta_R)$. The robot can sample from $p(\theta_H^{post})$ to compute the probability of safety, which is a distribution that incorporates the proposed CBP model. We equip a probabilistic long-term safe controller from the literature \cite{wang2022myopically} with this distribution $p(\theta_H^{post})$ to compute safe trajectories for the robot by sampling from the uncertain human intention instead of just from a stochastic dynamics model as described in the original paper.




\section{Simulation Environment}
\subsection{Score Function}
For our two-agent goal-reaching tasks, we find the following score function to be expressive enough to enable proactive robot goal selection: 
\begin{equation}
\begin{split}
    &s(\theta_H, \theta_R; \mathbf{o^{0:t}}, \theta_H^{prior}) = w_1||x_H^t - \theta_H||\\
    &-w_2||\theta_H - \theta_R||
    + w_3||\theta_H - \theta_H^{prior}||,
\end{split}
\label{eqn:score_fn}
\end{equation}
where $w_1,w_2,w_3\in\mathbb{R}$ are hyperparameters\footnote{The weights or the score function itself could be learned from data or adapted online, an exploration of which we leave to future work.}.
This score function captures the idea that the human is more likely to choose a goal close to their current state (term 1), farther from the robot's goal (term 2), and prefer to not changing their current goal (term 3). A different score function would need to be chosen for different tasks, but in general should encode how likely the human is to change their goal from $\theta_H^{prior}$ to $\theta_H$ given that the robot chooses $\theta_R$.

\subsection{Simulated Human Models}
\label{sec:h_goal}

\noindent\textbf{Uncertain Human:} The first simulated human infers the robot's goal online (a common paradigm from literature~\cite{dragan2013legibility, huang2019nonverbal, huang2019enabling}) by keeping a belief $b_H^t(\theta)$ over the robot's goal. It chooses the closest goal that is not the same as the robot's most likely goal once its belief is sufficiently low-entropy ($\max_{\theta}b_H^t(\theta) \geq 0.4$):
\begin{align}
\begin{split}
    &\theta_H^* = \argmin{\theta\in\Theta} || x_H^t - \theta ||, \;\: \theta \neq\argmax{\theta^{'}\in\Theta} b_H^t(\theta^{'}).
\end{split}
\end{align}


\noindent\textbf{Stubborn Human:} The second simulated human is a stubborn human that will never change their goal (also a mode of human behavior studied in prior literature~\cite{nikolaidis2017human, devin2017decisions}). This human chooses its goal to be the goal closest to itself 
\begin{equation}
    \theta_H^* = \argmin{\theta\in\Theta} || C_H x_H - \theta ||.
\end{equation}

\subsection{Baseline Robot Controllers}
\label{sec:baseline_robots}
\noindent\textbf{Naive Robot:} This robot simply chooses the closest goal:
\begin{equation}
    \theta_R^* = \argmin{\theta\in\Theta} ||C_R x_R - \theta||
\end{equation}

\noindent\textbf{Reactive Robot:} For a less naive baseline, we use a reactive goal selection method based on observational Bayesian inference, which picks a goal that is close to the robot but is not the inferred goal of the human:
\begin{equation}
\begin{split}
    \theta_R^* &= \argmin{\theta\in\Theta} ||C_R x_R - \theta||, \;\: \theta \neq \argmax{\theta'\in\Theta} b_R^t(\theta')
\end{split}
\end{equation}


\noindent\textbf{Proactive-NN Robot:} For a strong baseline, we implement a learning-based conditional behavior prediction model that has a similar structure to other learning-based CBP models from the literature~\cite{tolstaya2021identifying, salzmann2020trajectron++}. It takes in the agents' trajectory history and the robot's future plan and outputs the probability of the human reaching each goal. We construct a dataset of 3.5 million data points 
and use the same proactive goal-selection method \eqref{eqn:final_objective} for this approach.

\begin{table}
    \centering
    \begin{tabular}[width=\columnwidth]{@{}lcc@{}}
    \toprule
    & \% rollouts $H$ changed $\theta$  & \# times $H$ changed $\theta$  \\ \midrule 
    Naive & $61\%$ & $0.8\pm0.8$ \\ 
    Reactive & $42\%$ & $0.7\pm1.1$ \\ 
    Proactive-NN & $\mathbf{17\%}$ & $\mathbf{0.27\pm0.7}$ \\ 
    Proactive-Model & $\mathit{18\%}$  & $\mathit{0.31\pm0.8}$ \\
    \bottomrule
    \end{tabular}
    \caption{\label{tab:courtesy_metrics} Average courtesy metrics in simulation (mean $\pm$ SD).}
\end{table} 

\begin{figure}
    \centering
    \includegraphics[width=0.7\columnwidth]{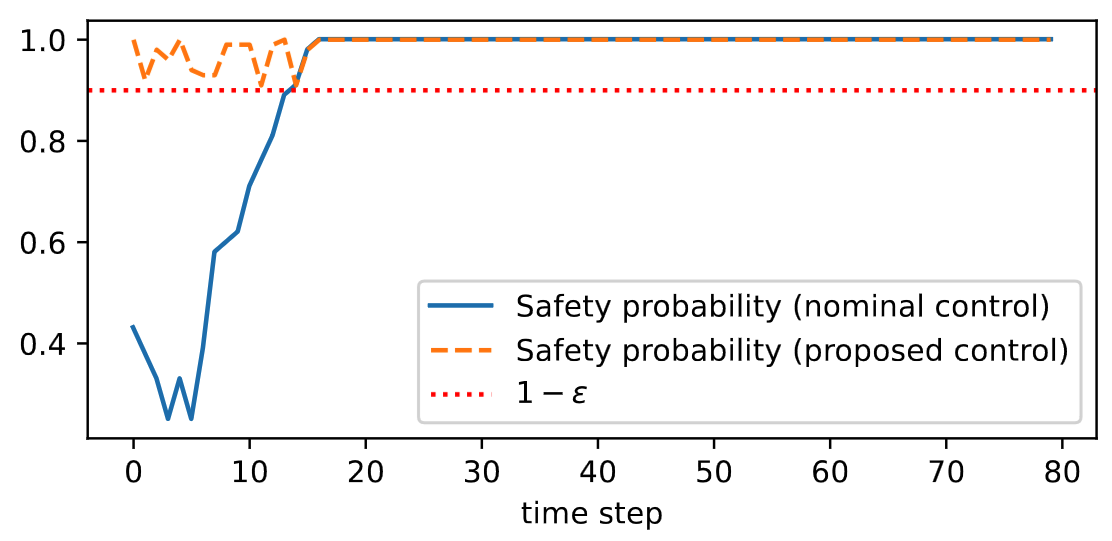}
    \caption{Long-term safety probabilities in simulation.}
    \label{fig:safe_probs}
    \vspace{-0.2in}
\end{figure}

\section{Simulation Results}
\label{sec:sim_results}

\begin{table}
    \centering
    \begin{tabular}[width=\columnwidth]{@{}lccc@{}}
    \toprule
    & stubborn human & uncertain human & overall   \\ \midrule 
    Naive & $3.2\pm3.1$ & $\mathbf{7.68\pm4.0}$ & $5.44\pm4.4$ \\ 
    Reactive & $6.4\pm3.0$ & $4.76\pm2.1$ & $5.6\pm2.7$ \\ 
    Proactive-NN & $\mathbf{9.1\pm3.9}$ & $6.28\pm3.8$ & $\mathbf{7.7\pm4.1}$ \\ 
    Proactive-Model & $\mathit{8.6\pm4.9}$ & $6.4\pm3.5$ & $\mathit{7.5\pm4.0}$ \\
    \bottomrule
    \end{tabular}
    \caption{\label{tab:sim_efficiency} Average number of goals reached in simulation (mean $\pm$ SD).}
    \vspace{-0.2in}
\end{table} 

We tested the proposed controller against a simulated human that is randomly selected to be either ``uncertain'' or ``stubborn,'' (\sref{sec:h_goal}), unknown to the robot, for 100 simulated games. Each game lasts for 30 seconds. For the proactive-model robot, the weights in \eqref{eqn:score_fn} are fixed at $[w_1,w_2,w_3]=[2,0.9,2]$ for both human models.

\noindent\textbf{Proactive models are courteous.} First, we test how courteous each robot agent is around the uncertain human (\tabref{tab:courtesy_metrics}). We measure the percentage of trajectories that the human changed their initial goal and the average number of times per trajectory that the human changed their goal. The proactive-NN controller performs the best\footnote{The NN-based controller is likely not near 0 because of distribution shift between training and testing.} with our proposed proactive-model controller barely behind it. The proactive-model controller is able to perform just as well as the learning-based baseline in a data-sparse environment by just doing online inference.

\noindent\textbf{Proactive models are flexible.} We measure the efficiency of the interactions by counting the total number of goals reached by the team (\tabref{tab:sim_efficiency}) over 100 initial conditions. The proactive-NN model performs the best on average with our proactive-model controller just behind it, showing us that the proactive models are flexible enough to efficiently interact with very different kinds of humans without prior knowledge. This also shows that the proposed controller is able to perform well in a data-sparse environment. We do, however, see that the naive controller works best with the uncertain human. Because the naive controller will immediately commit to an action, it's very easy for the simulated human to respond immediately, whereas the other controllers take longer because they first estimate the human's intent. 



\noindent\textbf{Long-term safety can be assured.} \figref{fig:safe_probs} shows the safety probability under time horizon $H=20$ and risk tolerance $\epsilon=0.1$ with and without the long-term safe controller while the robot selects goals with the proactive-model controller. We can see that with safety the modification, long-term safety probability is maintained over $90\%$ as desired while the nominal goal-seeking controller fails to do so.

\section{User Study}
\label{sec:user_study}
We ran an IRB-approved study with real users interacting with our 2D environment. The study lasted 20 minutes and participants were paid $\$5$.

\begin{figure}
    \centering
    \includegraphics[width=0.8\columnwidth]{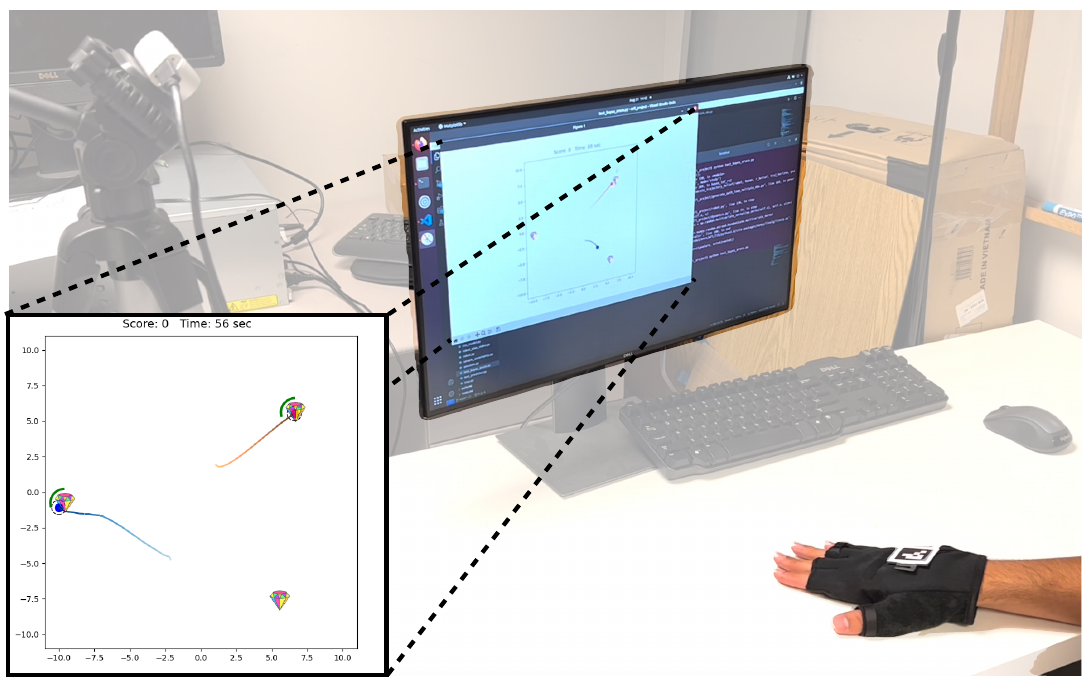}
    \caption{Shows the setup for the user study where participants control an on-screen avatar directly with their hand to try and collect diamonds in collaboration with different robots.}
    \label{fig:user_study_setup}
    \vspace{-0.25in}
\end{figure}

\noindent\textbf{Experimental Setup:} Users control an avatar on a computer screen with their hand. Their hand position is measured by a camera and mapped onto the 2D space (\figref{fig:user_study_setup}). The task is for the user to collect diamonds simultaneously with the (virtual) robot in the allotted time without colliding. Each participant interacted with all three robot types in a randomized order (four 45-second games per robot).

\noindent\textbf{Independent Variables:} We ran a within-subjects study and manipulated the \textit{robot type} with three levels: \textit{naive} (\sref{sec:baseline_robots}), \textit{reactive} (\sref{sec:baseline_robots}) and \textit{proactive} (\sref{sec:proposed_controller}). We do not include the proactive-NN robot due to the lack of a large dataset of real people doing this task.

\noindent\textbf{Objective Measures:} We cannot directly measure if the human's intention changed over time, so we instead measure the \textit{hesitation time}, or the amount of time the human waits before reaching for a goal, as a proxy for how much mental energy participants exert to choose their goal. 
We additionally measure the team's \textit{efficiency} as the number of goals reached.

\noindent\textbf{Subjective Measures:} We ask participants a series of 5-point Lkert scale~\cite{likert1932technique} questions about their experience with each robot and finally ask for a ranking of all robots. Here, we show the questions most relevant to the robot controller, but users were also asked general questions about their overall satisfaction and strategies on the task.

\noindent\textbf{Hypothesis H1:} \textit{Participants interacting with the proactive robot will hesitate less and score more points than when interacting with the reactive robot.}

\noindent\textbf{Hypothesis H2:} \textit{Participants will feel that the proactive robot was easier to work with because it better accounted for their intentions.}


\noindent\textbf{Participants:} We recruited 22 participants (largely with technical backgrounds) from the campus community.  One user's data was not included due to technical difficulties.

\begin{table}
    \centering
    \begin{tabular}[width=\columnwidth]{@{}lcc@{}}
    \toprule
    & \# goals reached & avg. hesitation time (\textit{sec}) \\ \midrule
    Naive & $\mathbf{18.9\pm2.1}$ & $ 0.95\pm0.34$ \\
    Reactive & $15.2\pm1.6$ & $1.2\pm0.55$\\
    Proactive-Model & $17.1\pm2.3$ & $\mathbf{0.87\pm0.37}$ \\
    \bottomrule
    \end{tabular}
    \caption{\label{tab:objective_results} Objective results for user study (mean $\pm$ SD).}
    \vspace{-0.1in}
\end{table} 

\begin{table}
    \centering
    \begin{tabular}[width=\columnwidth]{@{}lccc@{}}
    \toprule
    & Naive & Reactive & Proactive \\ \midrule 
    \textbf{Accounted: }``[Robot] accounted for \\ the [goal] I wanted to pick when\\it was picking a [goal].'' & $2.1$ & $\mathbf{4.0}$ & $3.7$ \\ \midrule
    \textbf{Changed: }``I often changed which [goal]\\ I picked initially because of [Robot].'' & 3.7 & \textbf{1.7} & 2.7 \\ \midrule
    \% ranked 1st (forced choice) & 19\% & 33\% & \textbf{47\%} \\ 
    \bottomrule
    \end{tabular}
    \caption{\label{tab:likert_qs} Responses to subjective survey questions (5: Strongly Agree, 1: Strongly Disagree).}
    \vspace{-0.25in}
\end{table}

\noindent\textbf{Quantitative Results:} Running one-way repeated measures ANOVAs tell us the robot type had a statistically significant effect on the number of goals reached by the team (\tabref{tab:objective_results}, $F(2, 40)=67.1, p<0.0001$) and on users' hesitation time ($F(2,40)=14.7, p < 0.0001$). A post-hoc Bonferroni test on the score tells us that the differences in goals reached between all three pairs are significant ($p<0.001$ for all). Users performed the best with the Naive robot, second best with the Proactive robot and worst with the Reactive robot. This partially supports \textbf{H1}, though we did not expect users to perform so well with the naive robot---users were ultimately capable of being followers, even if they did not necessarily enjoy it. A post-hoc Bonferroni test on the hesitation time tells us that users hesitated significantly less with the proactive robot than the reactive robot ($p=0.0004$) and less with the naive robot than the reactive robot ($p=0.004$). This result also supports \textbf{H1}, and we see a surprisingly positive result with the naive robot, likely meaning users were able to quickly react to the naive robot's goal selection. Throughout the interaction, users could likely easily predict the naive robot's actions, telling us that predictability is also important in designing efficient collaborative robots.

We also empirically measured the safety of the interactions to check that they line up with the theoretical guarantees. Across interactions with all 21 users, there were 13 collisions out of $45 \mathrm{s} \times 10 \mathrm{Hz} \times 4 \text{ trials} \times 21 \text{ users}=37800$ timesteps, for an empirical safety probability of $99.96\%$.

\noindent\textbf{Qualitative Results:} 
To test \textbf{H2}, we ran a repeated measures ANOVA on the effect of the robot type on the survey question ``The robot was easy to collaborate with,'' but found no significant differences ($F(2,40)=2.9, p=0.06$), since users rated all three as easy to collaborate with. We did, however, find significant differences between the robots on the ``Accounted'' (\tabref{tab:likert_qs}, $F(2, 40)=12.4,p<0.0001$) and ``Changed'' questions ($F(2,40)=18.5,p<0.0001$). A post-hoc Bonferroni test tells us that, relative to the naive robot, participants thought the proactive ($p=0.0006$) and reactive ($p=0.0005$) robots accounted more for their goal selection. This seems to support \textbf{H2}, although participants tended to think the proactive robot was about as responsive to their goal selection as the reactive robot. A post-hoc Bonferroni test tells us that participants felt that they changed their goals around the naive robot, while they did not change their goals around the proactive robot ($p=0.01$) or around the reactive robot ($p<0.0001$), also supporting \textbf{H2}. They also felt that they changed their goal less with the reactive robot than the proactive robot ($p=0.002$). 


We additionally look at users' rankings of the three robots (\tabref{tab:likert_qs}) to test \textbf{H2}. While not a majority, a plurality of participants (47\%) ranked the proactive robot as the most preferred interaction partner when asked to rank all three robots, which supports \textbf{H2}. Treating the participants' rankings as votes for the robots and running an instant-runoff-voting election shows that the proactive robot is the winner of the ``election.'' 

\section{Conclusion and Future Work}
We proposed a Bayesian-inference-based formulation of conditional behavior prediction that can allow a robot to reason about the affect its future actions will have on a human in collaborative multi-goal settings. The CBP formulation was based on the insight that we can decouple the human's prior goal selection $\theta_H^{prior}$ from their posterior goal selection $\theta_H^{post}$ that accounts for the robot's actions. We additionally discussed how to utilize a CBP model for long-term safety as well as for switching between courteous and influential behavior. In simulations, we found that our proposed controller allows the robot to be efficient at interacting with different kinds of humans (even without tuning the score function weights) and that it performs very similarly to a learning-based CBP model without the need for a large dataset. We tested our proposed controller in a user study and found that although users tended to perform the best with the naive baseline robot, users tended to \textit{prefer} interacting with our proactive controller. Future work may try to improve the model-based CBP approach by adapting the score function weights online, which would keep the data-efficiency benefits while potentially increasing the robot's efficiency at acting with different kinds of real human partners.


\section*{Acknowledgments}
This material is based upon work supported by the National Science Foundation (NSF) Graduate Research Fellowship under Grant No. DGE1745016 and DGE2140739 and additionally under Grant No. 2144489 as well as the Manufacturing Futures Institute, Carnegie Mellon University, through a grant from the Richard King Mellon Foundation. Any opinions, findings, and conclusions or recommendations expressed in this material are those of the authors and do not necessarily reflect the views of the NSF.

\bibliographystyle{IEEEtran}
\bibliography{references}


\appendices
\section{Human's Bayesian Goal Prediction}
\label{sec:bayesian}
The human assumes the robot is noisily rational in choosing its actions for an intended goal (or that the robot is exponentially more likely to choose an action if it has a higher Q-value):
\begin{equation}
    p(u_R^t \mid x_R^t; \theta) = \frac{e^{\beta_H Q(x_R^t,u_R^t;\theta)}}{\int_{u_R^{'}} e^{\beta_H Q(x_R^t,u_R^{'};\theta)} },
    \label{eqn:h_bayes_likelihood}
\end{equation}
where $\beta_H$ is the rationality coefficient (sometimes called the ``inverse temperature'' parameter or the ``model confidence''). In general, the integral in the denominator can be challenging to compute, but the simulated human makes use of the robot's baseline LQR controller to define the reward function as the instantaneous LQR cost:
\begin{equation}
    r_R(x_R,u_R;\theta) = -(x_R-\theta)^\top Q(x_R-\theta) - u_R^\top Ru_R,
\end{equation}
so the $Q-$function is then the negative optimal cost-to-go:
\begin{equation}
    Q_R(x_R, u_R; \theta) = r(x_R,u_R;\theta) - (x'-\theta)^\top P(x'-\theta)
\end{equation}
where $x'=Ax_R+Bu_R$ and $P$ is the solution to the discrete-time algebraic Ricatti equation (DARE): 
\begin{equation}
    P=A^\top P A-A^\top P B(R+B^\top PB)^{-1}B^\top PA+Q.
\end{equation}

Following prior work \cite{tian2023towards}, this allows us to compute an exact form of the denominator of \eqref{eqn:h_bayes_likelihood}:
\begin{equation}
\begin{split}
    &\int e^{\beta_R Q_R(x_R,u_R;\theta)} \\
    &= e^{-\beta_R(x_R-\theta)^\top P(x_R-\theta)}\sqrt{\frac{(2\pi)^m}{\text{det}(2\beta_R R+2\beta_RB^\top PB)}}.
\end{split}
\end{equation}

The human then uses this likelihood function (and Bayes Rule) to update its belief given a new observation $(x_R^t,u_R^t)$:
\begin{equation}
\begin{split}
    b_H^t(\theta_i) &= p(\theta_i \mid x_R^{0:t}, u_R^{0:t})\\
    &= \frac{p(u_R^{t} \mid x_R^{t}; \theta_i) p(\theta_i \mid x_R^{0:t-1}, u_R^{0:t-1})}{\sum_{\theta^{'}} p(u_R^{t} \mid x_R^{t}; \theta^{'}) p(\theta^{'} \mid x_R^{0:t-1}, u_R^{0:t-1})}
\end{split}
\end{equation}

\section{Learning-based CBP Details}
The basic form of these models is that they take in a trajectory history for all agents to as well as the \textit{future trajectory} of the robot and outputs a prediction about the non-ego agents (the human in our case). We structure our learning-based model similarly, it takes in the human and robot past trajectories $x_H^{t-k:t}, x_R^{t-k:t}$ as well as the robot's \textit{future plan} $x_R^{t+1:t+T}$. The trajectories are fed into LSTM layers with attention, and finally all pieces are concatenated with the set of goals $\{\theta_1,\ldots,\theta_N\}$ and passed into linear layers to output the probability of the human reaching each goal $[P(\theta_1),\ldots,P(\theta_N)]$. The network is trained to output the empirical probability of the human reaching each potential goal conditioned on the robot's plan; it is trained with an MSE loss.

These papers have generally been in spaces where there are large pre-existing datasets to train on, like in social navigation and autonomous driving. We're instead focusing on human-robot collaborative tasks such as manufacturing and in-home robots, so such datasets are hard to come by. As a result, we create our own dataset for our simulated human-robot collaboration task. The dataset consists of 3.5 million data points, which corresponds to approximately 97 hours of data collected for this particular human-robot interaction. This kind of data would be impractical to collect on a real human-robot interaction, so we use this as a baseline in our simulations but not in our user studies. We use the same goal-selection method \eqref{eqn:final_objective} for this approach as our proposed controller, which works since this method is also a conditional prediction method.

\begin{table*}
    \centering
    \begin{tabular}[width=\columnwidth]{@{}lccccc@{}}
    \toprule
    & Naive & Reactive & Proactive & F-score & p-value \\ \midrule 
    ``I felt like the robot accounted for the [goal] I wanted to pick when\\it was picking a [goal].'' & $2.1$ & $\mathbf{4.0}$ & $3.7$ & $F(2,40)=12.4$ & $p<0.01$\\ \midrule
    ``I often changed which [goal] I picked initially because of the robot.'' & 3.7 & \textbf{1.7} & 2.7 & $F(2,40)=18.5$ & $p<0.01$\\ \midrule
    ``I felt like the robot and I scored as many points as we could.'' & 3.6 & \textbf{4.1} & 3.0 & $F(2,40)=4.0$ & $p=0.02$\\ \midrule
    ``The robot influenced me to pick good [goals] for the team.'' & 3.0 & 2.6 & 3.2 & $F(2,40)=1.8$ & $p=0.18$\\ \midrule
    ``The robot's choice of [goals] made me choose worse [goals] for the team.'' & 3.3 & \textbf{1.8} & 2.5 & $F(2,40)=10.0$ & $p<0.01$\\ \midrule
    ``The robot was easy to collaborate with.'' & 3.3 & 4.1 & 3.4 & $F(2,40)=2.9$ & $p=0.06$\\ \midrule
    ``I felt like the robot picked the best [goals] to grab for the team.'' & 3.2 & \textbf{4.0} & 2.7 & $F(2,40)=8.1$ & $p<0.01$\\ \midrule
    ``I felt like the robot hindered the team's performance.'' & 2.8 & 2.2 & 3.0 & $F(2,40)=2.6$ & $p=0.08$\\
    \bottomrule
    \end{tabular}
    \caption{\label{tab:full_likert_qs} Responses to subjective survey questions (5: Strongly Agree, 1: Strongly Disagree).}
\end{table*} 

\section{Safe Trajectory Generation}

\begin{algorithm*}
\caption{Safe Robot Control Pipeline with Model-Based CBP}\label{alg:model_based_cbp}
\begin{algorithmic}
\State \textbf{Given:} $\{G_1,...,G_N\}$ \Comment{goal locations}
\State \textbf{Given:} $x_H^0, x_R^0$ \Comment{agent starting positions}
\State $b_R^0(\theta_H^{prior}) \gets \text{uniform}$ $\hat b_H^0(\theta_R) \gets \text{uniform}$ \Comment{initial robot belief and mental model}
\While{$t < H$} \Comment{$H$ is the trajectory horizon}
    \State $b_R^t(\theta_H^{prior}) \gets (x_H^t, x_R^t, u_H^t)$, $\hat b_H^t(\theta_R) \gets (x_R^t, u_R^t)$ \Comment{robot updates nominal belief and mental model}
    \State $b_R^t(\theta_H^{post} \mid \theta_R) \gets \sum_{\theta_H^{prior}} p(\theta_H^{post} \mid \theta_H^{prior}, x_H^{0:t}, x_R^{0:t}, u_H^{0:t}, \theta_R) b^t_R(\theta_H^{prior})$ \Comment{CBP belief update}
    \State $p(\theta_H^{post})=\hat b_H^t(\theta_R) b_R^t(\theta_H^{post} \mid \theta_R)$ \Comment{robot computes overall goal probabilities}
    \State $\mathcal{X}_R \gets trajGen(x_H^t, x_R^t, p(\theta_H^{post}))$ \Comment{generate candidate safe trajectories}
    \State $\mathbf{x}_R^*, \theta_R^* \gets \argmin{(\mathbf{x}_R, \theta_R) \in \mathcal{X}_R} J(\mathbf{x}_R,b_R^t(\theta_H^{post} \mid \theta_R), b_R^t(\theta_H^{prior}))$ \Comment{choose (traj, goal) pair that minimizes cost}
    \State $u_R^t \gets \mathbf{x}_R^{u,0}$ \Comment{choose first control action from trajectory}
    \State $x_R^{t+1} \gets f_R(x_R^t, u_R^t)$
\EndWhile
\end{algorithmic}
\end{algorithm*}


We propose a long-term safe controller to ensure the probability of safety over the time horizon $H$ is always above the desired threshold $1-\epsilon$, as stated in the objective~\eqref{eq:safety_objective}.
The controller is summarized in Algorithm~\ref{alg:long_term_safe_control}. The key procedures at each time step are: 1. estimate the probability of safety for the current state 2. find the closest state that is safe enough if the initial state is not 3. execute goal-pursing control plus a potential field safe control that drives the robot to safe regions. Note that when estimating the safety probability, we assume a greedy human that has no robot-avoidance control to cover the worst case. 

\begin{remark}
    Assuming the human intention prediction $p(\theta_H^{post})$ is correct, the proposed long-term safe control (Algorithm~\ref{alg:long_term_safe_control}) is guaranteed to meet the design objective~\eqref{eq:safety_objective}. Similar theoretical derivations can be found in~\cite{wang2022myopically,wei2019safe}.
\end{remark}

To achieve more efficient real-time implementation, we use Gaussian distribution to approximate the uncertainty in human's motion at each time step $t$. 
Specifically, the mean of the Gaussian is the expected nominal motion of the human $x_H$ and the variance will be some estimated value $\sigma^2$.
For trajectory generation, we are interested in the long-term safety of the human-robot interaction in the sense that we want the generated trajectories to be safe at all time within a horizon $H$. Given the initial level of uncertainty $\sigma^{(i)}$ for each human mode $i$, we know that the uncertainty at time step $t$ is $\sigma^{t(i)} = \sqrt{t} \sigma^{(i)}$. 
For each possible goal $\theta_H^{(i)}$ of human, let $x_H^{0:H(i)}$ be the nominal trajectory of the human starting from $x_H^0$ pursuing $\theta_H^{(i)}$. The probability of human choosing $\theta_H^{(i)}$ is given by $\theta_H^{(i) {post}}$. The goal is to generate $x_R^{0:H}$ such that the following safety condition is satisfied
\begin{equation}
    \phi_{d_\text{min}+\sigma^{t(i)}}(x^t) \geq 0, \forall t < H, \; \forall i.
\end{equation}

For robot trajectory generation, we introduce the potential field controller to generate candidate trajectories that are more likely to be safe. The idea of potential field controller is that it will impose repelling forces to the controlled agent if it is close to the unsafe region. For example, if we want the robot's position $x_R$ to be away from the human's position $x_H$, we will have the potential field control for the robot to be
\begin{equation}
\label{eq:potential_field_control}
    u_{\text{pf}} = \frac{\gamma}{d^2}(C_H x_H - C_R x_R) = K_{\text{pf}}(x_H-x_R),
\end{equation}
where $d$ is the distance between human and robot and $\gamma$ is the repel force of the potential field controller. For simplicity we assume $C_H = C_R$ and use $K_{\text{pf}}$ to denote the coefficient. 
Similarly, we can define $o$ to be some obstacles to be avoided and use~\eqref{eq:potential_field_control} to find the safe control by replacing $x_H$ with $o$.

The overall safe trajectory generation algorithm in shown in Algorithm~\ref{alg:traj_gen}, where we use potential field controller to avoid different possible human motions given different goals, and use synthetic obstacles in the state space to diverse the trajectory. Safety is identified at each time step $t$ with uncertainty level $\sigma\sqrt{t}$, which characterizes the growing uncertainty bound over time. Experiments show that synthetic obstacles close to the initial position of the robot will give more diverse trajectories.


\begin{algorithm*}
\caption{Long-term Safe Control}\label{alg:long_term_safe_control}
\begin{algorithmic}



\Procedure{safeControl}{$x_H^t, x_R^t, \theta_R, \theta_H, H$}

\State $F \gets$ long\_term\_safe\_prob $(x_H^t, x_R^t, \theta_H, H)$ 

\If {$F > 1-\epsilon$}

\State $u_R^t \gets K(x_R^t - \theta_R)$ \Comment{goal pursuing control}

\Else

\State $x_{\text{safe}} \gets$ find\_safe\_state $(x_H^0, x_R^0, \theta_H, H, \epsilon)$

\State $u_R^t \gets K(x_R^t - \theta_R) - K_{\text{pf}} (x_R^t - x_{\text{safe}})$ \Comment{potential field safe control} 

\EndIf

\State \textbf{return} $u_R^t$
\Comment{return trajectories}
\EndProcedure

\Procedure{longTermSafeProb}{$x_H^0, x_R^0, \theta_R, \theta_H, H$}
\State \textbf{Given:} $n$ \Comment{number of episode to estimate safety probability}

\State \textbf{Initialize} $\{F^{(1)}, F^{(2)}, \cdots, F^{(n)}\} = \mathbf{1}$ \Comment{record safety of each trajectory}

\For {$k = 1:n$}
\State $\theta_H^* \gets \text{sample} (\theta_H) \text{ from } p(\theta_H^{post})$
\For {$t = 1:H$}

\State $u_R^t \gets K(x_R^t - \theta_R)$ \Comment{goal pursuing control for robot}
\State $u_H^t \gets K(x_H^t - \theta_H^*)$ \Comment{goal pursuing control for human}

\State $x_R^{t+1} \gets \text{step}(x_R^t, u_R^t)$

\State $x_H^{t+1} \gets \text{step}(x_H^t, u_H^t)$ \Comment{human dynamics with disturbance}

\If{$\|x_R^t - x_H^t\| < d_\text{min}$}
\State $F^{(k)} = 0$
\State \textbf{break}

\EndIf

\EndFor
\EndFor
\State \textbf{return} $\text{mean}(F)$
\EndProcedure

\Procedure{findSafeState}{$x_H^0, x_R^0, \theta_R, \theta_H, H, \epsilon$}
\State \textbf{Given:} $n$ \Comment{number of sampling states for a certain radius}

\State \textbf{Initialize} $r = 1$, $s = 0$ \Comment{searching radius and indicator of safe state found}

\While{$s = 0$}
\For {$k = 1:n$}
\State $x_R' \gets$ uniform\_sample $(x_R, r, k)$ \Comment{sample near robot's position with radius $r$}

\State $F \gets$ long\_term\_safe\_prob $(x_H^0, x_R', \theta_H, H)$ 

\If {$F > 1-\epsilon$}
\State $s = 1$ \Comment{mark safe state found}
\State \textbf{return} $\{x_R', F\}$ \Comment{return the safe state and its safety probability}
\EndIf
\EndFor
\EndWhile
\EndProcedure

\end{algorithmic}
\end{algorithm*}

\begin{algorithm*}
\caption{Candidate Trajectory Generation}\label{alg:traj_gen}
\begin{algorithmic}
\Procedure{trajGen}{$x_H^0, x_R^0, \theta_R, \theta_H^{prior}$}
    \State \textbf{Given:} $\{\sigma^{(1)}, \sigma^{(2)}, ...,\sigma^{(N)}\}$ \Comment{uncertainties in each mode of the human}
    \State $\{o_1, o_2, ..., o_M\} \gets$ sample\_obstacle$(x_R)$ \Comment{generate synthetic obstacles}
    \State \textbf{Initialize} $\{s_1, s_2, ..., s_M\} = \mathbf{1}$ \Comment{record safety of each trajectory}
    \For {$o_m \text{ in } \{o_1, o_2, ..., o_M\}$} \Comment{loop over all synthetic obstacles}
    \For {$t = 1:H$}
    \State $u_R^t \gets K(x_R^t - \theta_R)$ \Comment{goal pursuing control}
    \State $u_R^t \gets u_R^t + K_{\text{pf}} (x_R^t - o_m)$ \Comment{potential field control against obstacle}
    \For {$\theta_H^{(i)} \text{ in } \{\theta_H^{(1)},...,\theta_H^{(N)}\}$} \Comment{loop over all human's goals}
    \If {$\|x_R^t-x_H^{t(i)}\|_2 \leq d_\text{min}+\sigma^{(i)} \sqrt{t}$} \Comment{check safety}
    \State $s_i \gets 0$
    \State \textbf{break}
    \Else
    \State $u_R^t \gets u_R^t + \theta_H^{(i) prior} K_{\text{pf}} (x_R^t - x_H^{t(i)})$   \Comment{potential field control against each human mode}
    \State $u_H^{t(i)} \gets K(x_H^{t(i)} - \theta_H^{(i)})$ \Comment{goal pursuing control for human}
    \State $x_H^{t+1(i)} \gets \text{step}(x_H^{t(i)}, u_H^{t(i)})$
    \EndIf
    \EndFor
    \State $x_R^{t+1} \gets \text{step}(x_R^t, u_R^t)$
    \EndFor
    \If {$s_i = 1$}
    \State $\{X_R, U_R\} \gets [\{X_R, U_R\}, \{x_R, u_R\}]$ \Comment{record safe trajectory}
    \EndIf
    \EndFor
    \State \textbf{return} $\{X_R, U_R\}$
    \Comment{return trajectories}
\EndProcedure
\end{algorithmic}
\end{algorithm*}

\section{Subjective Questions for User Study}
\label{sec:full_user_study}
We ran a one-way repeated measures ANOVA for the effect of the robot type on each survey question listed in \tabref{tab:full_likert_qs} and ran post-hoc tests with Bonferroni tests. As noted in the main text, we found significant differences for the ``Accounted'' and ``Changed'' questions. We additionally found significant differences for the questions about scoring as many points as they could, influencing negatively, and picking the best goals for the team. Participants rated that they felt like the team scored as many points as possible more with the reactive robot than the proactive robot ($p<0.01$), which is interesting given that participants statistically performed the \textit{worst} with the reactive robot. Participants also rated that they choose worse goals with the naive robot than the reactive robot ($p<0.01$) and thought both the CBP and reactive robots were not making them choose worse goals. They additionally felt like the reactive robot picked better goals for the team than the CBP robot ($p<0.01$), which is again interesting because it does not align with the actual performance on the task. This seems to indicate that the peoples' subjective opinions of a collaborator may not be based precisely on performance, but on other social factors such as perceived intelligence or predictability. We also see that on these subjective questions, the reactive robot is most often the highest rated, but when forced to choose a preferred collaborator, they tended to choose the CBP model. This may be because they felt it was the most capable collaborator overall, even though they felt that the reactive robot was responding well to their strategies.

\end{document}